\documentclass{article}

\usepackage{PRIMEarxiv}

\usepackage[utf8]{inputenc} 
\usepackage[T1]{fontenc}    
\usepackage{hyperref}       
\usepackage{url}            
\usepackage{booktabs}       
\usepackage{amsfonts}       
\usepackage{nicefrac}       
\usepackage{microtype}      
\usepackage{lipsum}
\usepackage{fancyhdr}       
\usepackage{graphicx}       
\usepackage{comment}

\usepackage{lscape}
\usepackage{bm}
\usepackage{color}
\usepackage{ulem}
\usepackage{algorithm}
\usepackage{algorithmicx}
\usepackage{algpseudocode}
\usepackage{multicol}
\usepackage{amsmath,amssymb}

\newcommand{\ForAny}{\overset{\forall}{}}
\newcommand{\ForSome}{\overset{\exists}{}}

\newcommand{\anomaly}[1]{{\textit{#1}}}

\newcommand{\function}[3]{{#1}\colon {#2}\to {#3}}
\newcommand{\NaturalNumber}{\mathbb{N}}

\newcommand{\oVec}[1]{\bm{#1}}
\newcommand{\oMat}[1]{\mathbf{#1}}

\title{Long-term Detection System for Six Kinds of Abnormal Behavior of the Elderly Living Alone*}
\author{
  Kai Tanaka, Mineichi Kudo, Keigo Kimura, Atsuyoshi Nakamura \\
  Graduate School of Information Science and Technology \\
  Hokkaido University\\
  Sapporo, 060-0814, JAPAN\\
  \texttt{\{tanakai, kimura5, atsu\}@ist.hokudai.ac.jp}\\
  \texttt{mineichi@mbm.nifty.com} 
}

\begin{document}
\maketitle

\begin{abstract}
The proportion of elderly people is increasing worldwide, particularly those living alone in Japan. As elderly people get older, their risks of physical disabilities and health issues increase. To automatically discover these issues at a low cost in daily life, sensor-based detection in a smart home is promising. As part of the effort towards early detection of abnormal behaviors, we propose a simulator-based detection systems for six typical anomalies: being semi-bedridden, being housebound, forgetting, wandering, fall while walking and fall while standing. Our detection system can be customized for various room layout, sensor arrangement and resident's characteristics by training detection classifiers using the simulator with the parameters fitted to individual cases. Considering that the six anomalies that our system detects have various occurrence durations, such as being housebound for weeks or lying still for seconds after a fall, the detection classifiers of our system produce anomaly labels depending on each anomaly's occurrence duration, e.g., housebound per day and falls per second. We propose a method that standardizes the processing of sensor data, and uses a simple detection approach. Although the validity depends on the realism of the simulation, numerical evaluations using sensor data that includes a variety of resident behavior patterns over nine years as test data show that (1) the methods for detecting wandering and falls are comparable to previous methods, and (2) the methods for detecting being semi-bedridden, being housebound, and forgetting achieve a sensitivity of over 0.9 with fewer than one false alarm every 50 days.
\end{abstract}


\thanks{*A part of this paper was published in \cite{Tanaka2024FallDetectionby}. This paper is submitted to a journal for review on November 6, 2024.  This work was supported by JST SPRING, Grant Number JPMJSP2119.}

\newcommand{\myfig}[4][width=120mm]{
  \begin{figure}[tbph]
    \centering
    \includegraphics[#1]{#2}
    \caption{#3}
    \label{#4}
\end{figure}
}

\section{Introduction}
The global population is aging. As a result, the share of the elderly aged 65 years or older in the world was 9.3\% (727 million persons) in 2020 \cite{UnitedNations2020WorldPopulationAgeing}. And the elderly living alone also increases mainly in developed countries. For instance, in 20 countries, including France and the United Kingdom, the percentages of elderly persons who live alone exceed 30\% between 2006 and 2015 \cite{He2016AnAgingWorld}. As the most aging country, Japan's share of the elderly population is 26.6\% in 2015 \cite{He2016AnAgingWorld}, and about 18\% of them live alone \cite{UnitedNations2020WorldPopulationAgeing}.

As persons get older, they become vulnerable to disabilities including physical and cognitive issues. For example, 27.7\% of elderly people (aged 65 and older) in the United States in 2019 have a mobility disability. The mini-mental state examination (MMSE) \cite{Folstein1975Mini-mentalstate}, a numerical evaluation scale of cognitive function, also decreases by a median of 4 points from the 18-24 age group to the 80+ age group \cite{Crum1993Population-basednormsfor}. This scale is known to be related to dementia \cite{Barry2010StagingDementia} and anomalies, including falls \cite{Anstey2006An8-yearprospective} and wandering \cite{Algase2009EmpiricalDerivationand, Algase2009Newparametersfor}.

To maximize the healthy life expectancy and the dignity of independence, continuous monitoring to watch the behavior of elderly persons living alone in their house is required. However, many developed countries face a lack of long-term care workers \cite{Hussein2005Aninternationalreview} especially in Japan \cite{Kondo2019Impactofincreased}.

Smart homes can be an alternative to human labor. Smart homes are homes with installed sensors used to monitor the resident's health and behaviors \cite{Cook2012HowSmartIs}. Smart homes have been practically tested in research with large testbeds since the 2010s \cite{Kaye2011Intelligentsystemsfor, Skubic2012Testingclassifiersfor, Cook2013CASAS:ASmart, Doyle2014Anintegratedhome-based}. To assess residents' health continuously and long-term while preventing physical burden and preserving their privacy, sensors in smart homes are often chosen from the viewpoint of their ubiquity and unobtrusiveness \cite{Rashidi2013ASurveyon, Kaye2011Intelligentsystemsfor}. In particular, ambient sensors such as motion infrared sensors and pressure sensors have advantages in this manner more effectively than cameras and wearable sensors \cite{Deep2019Asurveyon, Vallabh2018Falldetectionmonitoring}. Some studies utilized ambient sensors to monitor the long-term cognitive status \cite{Hayes2008Unobtrusiveassessmentof,Alberdi2018SmartHome-BasedPrediction} and anomalous behavioral deviation \cite{Lundstrom2016Detectingandexploring, Shin2011Detectionofabnormalliving, Forkan2015Acontext-awareapproach, Saives2015Activitydiscoveryand}. However, mere detection of rare events or outliers may indicate disruptions in daily rhythms, which are not always important. When an anomaly is detected, users need to review the data and manually verify the symptoms or causes. It is more useful to detect significant anomalous behaviors and well-known risk, such as falls or being housebound.

The ultimate objective of this study is to develop algorithms to detect typical anomalies of elderly individuals living alone in a smart home over a long period. For now, we select six typical anomalies: falls while walking, falls while standing, wandering, forgetting to turn off the home appliances and faucets, being housebound, and being semi-bedridden as shown in Tab. \ref{tab:six_implemented_anomalies}.

There are studies to detect typical anomalies including fall \cite{Chaccour2015Smartcarpetusing, Rimminen2010Detectionoffalls, Alwan2006Asmartand, Yazar2013Falldetectionusing, Popescu2012VAMPIR-anautomaticfall, Tao2012Privacy-preservedbehavioranalysis} and wandering \cite{Oliveira2022CNNforElderly, Lin2018Detectingdementia-relatedwandering, Ota2011Elderly-caremotionsensor, Zhao2014Alight-weightsystem, Gochoo2017Device-freenon-privacyinvasive, Chaudhary2020Sensorsignals-basedearly, Khodabandehloo2020Collaborativetrajectorymining}. There are also studies of sensor systems to detect forgetting \cite{Logeshwaran2022DesigninganIoT, Hsu2019ApplicationofInternet, Wai2011Pervasiveintelligencesystem, Doyle2014Anintegratedhome-based}. In related to being semi-bedridden and housebound, residents' daily activities, such as sleep \cite{Shahid2022Recognizinglong-termsleep} and outing patterns \cite{Tominaga2012Aunifiedframework}, are modeled. Anomalous sleep patterns have been detected \cite{Forkan2015Acontext-awareapproach, Yahaya2019AConsensusNovelty}. However, these studies only focus on each anomaly individually and do not address multiple anomalies with different occurrence durations, such as seconds-wise anomalies like falls, hours-wise anomalies like forgetting, and weeks-wise anomalies like being housebound. Simply combining these detection methods does not necessarily ensure efficient detection of multiple anomalies. Additionally, it is not cost efficient to prepare different types of sensors and sensor placements for each anomaly. Furthermore, the processing of sensor data is not standardized, making it difficult to address anomaly detection within a common framework.

In this paper, we propose a simulator-based system that detects the six types of typical anomalies with different occurrence durations at once in real time, using the data from a set of ambient sensors such as infrared motion sensors, pressure sensors, door sensors, flow sensors, and power sensors (as shown in Tab. \ref{tab:list_of_ambient_sensors}). To train the detection classifiers, our system uses the data that is generated by a simulator with parameters fitted to individual cases, instead of real data in long-term date, which is difficult to collect for three reasons: (1) it requires significant time and cost including running costs and participation fees, (2) anomalies are uncertain to occur, and (3) sensor placement is fixed and multiple arrangements cannot be tested. According to the experimental results using nine-years simulation training and test data including six types of anomalies, our system show usable performance.

The contributions of this study are as follows:
\begin{itemize}
  \item\textbf{C1.} Our detection system can be customized for various floor plans, sensor arrangement, resindent’s activity patterns, and anomaly patterns by setting simulator parameters to the values appropriate for each case. Anomaly patterns can be implemented in the simulator such as abnormal activity statistics and abnormal walking trajectories. This approach is promising because it does not rely on real smart home data, which is challenging to collect.
  \item\textbf{C2.} We propose a method to detect six types of abnormal behaviors simultaneously, using a unified sensor preprocessing framework from ambient sensors, which is unobtrusive such as infrared motion sensors, pressure sensors, door sensors, power sensors and flow sensors. This is cost effective to handle various occurrence durations of different anomalies, e.g., detecting falls that occur over several seconds and detecting being housebound that occur over weeks.
  \item\textbf{C3.} We propose the novel detection methods for forgetting to turn off appliances, being housebound, and being semi-bedridden, for which specific detection methods with ambient sensors have not been adequately developed and evaluated. Additionally, we also detect three other anomalies: fall while walking, fall while standing and wandering.
  \item\textbf{C4.} The proposed detection methods show practically usable performance for simulated sensor test data with anomaly generated using predefined anomaly models. In a performance metric that considered anomalies as correctly detected if the detected period overlaps even slightly with the true abnormal period, the sensitivity and false alarm rate per day are as follows: being semi-bedridden: 1.0 and 0, being housebound: 1.0 and 0.004, forgetting: 1.0 and 0.01, wandering: 1.0 and 0.017, fall while walking: 0.75 and 0.02, and fall while standing: 0.92 and 0.
\end{itemize}

\begin{figure}[tbp]
\centering
\includegraphics[width=120mm]{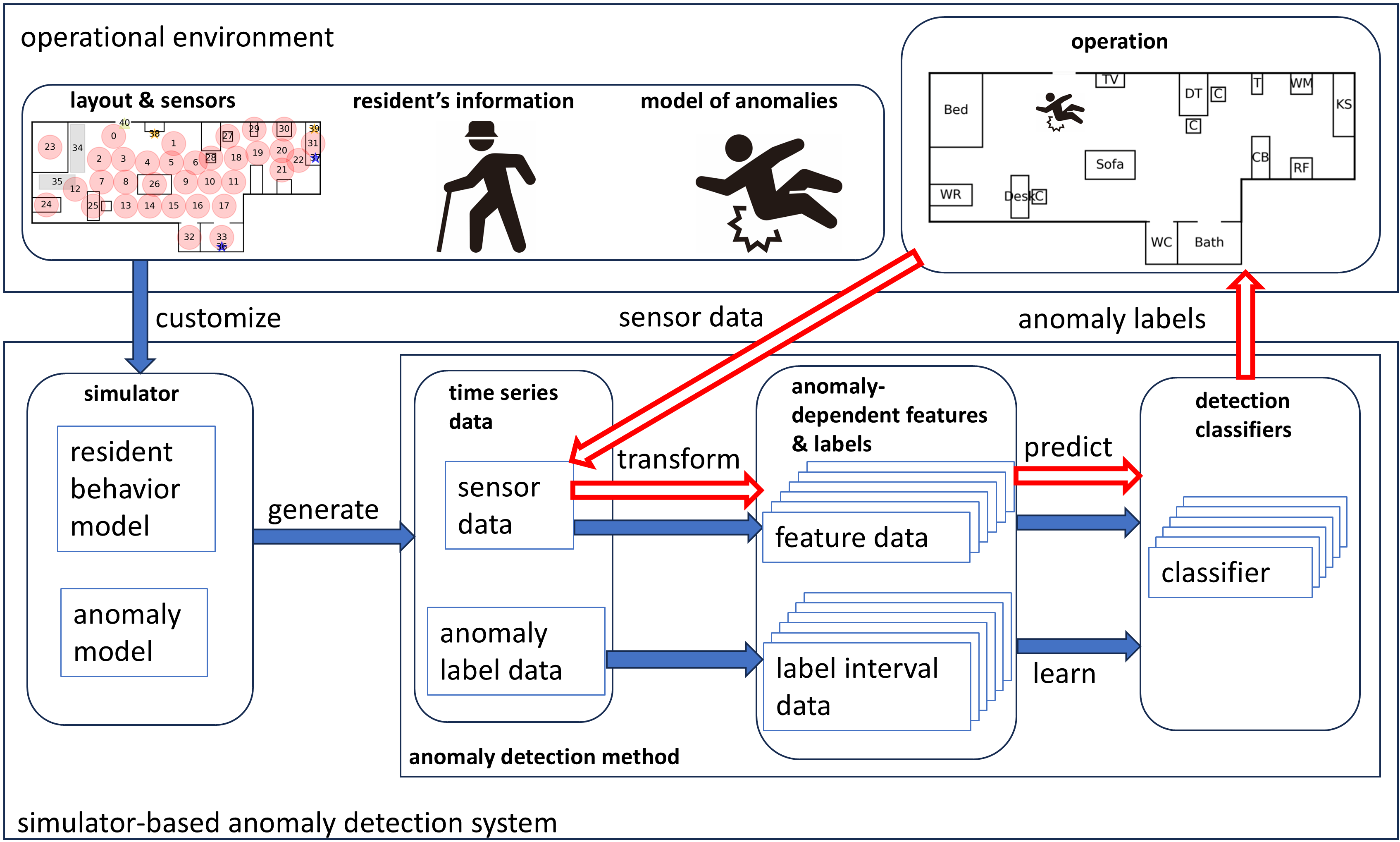}
\caption{Overview of systems. Anomaly detection classifiers are learned by the simulation data generated from simulators customized based on the information of operational environment. The simulated time series data are transformed into anomaly-dependent features and label interval data to apply different types and durations of anomalies.}
\label{fig:system_overview}
\end{figure}

\begin{table*}
\centering
\caption{Implemented six anomalies.}
\label{tab:six_implemented_anomalies}
\begin{tabular}[tbp]{p{2.5cm}p{3.3cm}p{4.5cm}p{6cm}}
\\
\hline
Anomaly & Symptom & Causes & Implemented modeling\\ \hline
\anomaly{Being semi-bedridden} & The resident often sleeps and avoids movement, indicating a status of grade 2 or 3 in an Eastern Cooperative Oncology Group (ECOG) performance \cite{Oken1982Toxicityandresponse}. & Bedridden is preceded by various kinds of disabilities due to health problem such as stroke, weakened body by aging and injuries caused by falls \cite{Normala2020BedriddenElderly_Factors}. In addition, housebound is considered as a risk of becoming bedridden \cite{Ishikawa2006Factorsrelatedto}. & The average frequency per month is $1/20$ and average duration is 30 days. During this period, the resident's daily activities change statistically, with an increase in the average duration of nap and rest activities and a decrease in both the average frequency and duration of going-out activities.\\
\anomaly{Being housebound} & The resident is confined to their home more than one week 
      \cite{Ganguli1996Characteristicsofrural}. & There is a correlation between the occurrence of being housebound and dementia, but causal relationship is unknown \cite{Lindesay1993Houseboundelderlypeople}. & The average frequency per month is $1/10$ and average duration is 14 days. During this period, the resident's daily activities statistically change, with a decrease in both the average frequency and duration of going-out and phone-calling activities.\\
\anomaly{Indoor wandering} & The resident walks around with anomalous travel patterns including randomly roundabout by the Martino-Saltzman model \cite{Martino-Saltzman1991TravelBehaviorof}. & Mini-mental state examination (MMSE) score \cite{Folstein1975Mini-mentalstate} is known to be effective to connect the patient to the degree of wandering \cite{Algase2009EmpiricalDerivationand}. Especially, the duration time is reported to be significantly associated with the MMSE score \cite{Algase2009Newparametersfor}. & $^{*}$The average frequency per month is $-1.86M + 56$ and average duration $-0.31M+9.8$ minutes. Wandering is defined as continuous walking by the resident without any specific intention, with random staging points selected from activity areas.\\
\anomaly{Forgetting} & The resident forgets to turn off home appliances and faucets. & Age and frequency of forgetfulness are significantly correlated \cite{Szabo2011Cardiorespiratoryfitness_hippocampal}. & $^{*}$The average frequency per month is $-M + 30$. Forgetting occurs randomly after the resident uses a home appliance, and the appliance is safely turned off when the resident returns near to the area. As a result, the duration varies over time.\\
\anomaly{Fall while walking} & The resident suddenly falls down while he or she is walking \cite{Chaccour2016Fromfalldetection}. & The longitudinal associations between MMSE and the falling risk are significantly strong at the individual level for predicting, and cognitive degeneration are predictive of falling risk\cite{Anstey2006An8-yearprospective}. & $^{*}$The average frequency per month is $-M/15+2$. The resident falls at a random middle point during a walk and remains lying down without moving for an average of 30 seconds, with an increased body range. Afterwards, the resident resumes walking. We define the fall as the period from the fall to the restart of walking.\\
\anomaly{Fall while standing} & The resident loses their balance when he or she starts standing up from sitting, or vice versa \cite{Nyberg1995Patientfallsin}. & Same with \anomaly{fall while walking}. & The model is almost the same as \anomaly{fall while walking}, but the fall location is limited to where the resident starts moving or reaches the bed. \\
\hline
\end{tabular}
\\
\flushleft{$^{*}$: $M$ is the MMSE simulated by the simple autoregressive model \cite{Tanaka2024SensorDataSimulation}.}
\end{table*}

\section{Related works}

Many studies use ambient sensors such as infrared motion sensors and pressure sensors to detect abnormal indoor behaviors of residents. Similar to our purpose, some studies aim to detect typical anomalies such as falls and wandering, as listed in Tab. \ref{tab:related_works_1} and Tab. \ref{tab:related_works_2}.

Falls are typical anomalies of the elderly. In a 1988 study \cite{Blake1988FallsbyElderly}, 35\% of individuals aged 65 and older fell at least once a year. Because falls are well-known risks for elderly individuals in indoor activities \cite{Yared2016Ambienttechnologyto}, there are many studies for detecting falls while walking \cite{Yazar2013Falldetectionusing, Popescu2012VAMPIR-anautomaticfall, Tao2012Privacy-preservedbehavioranalysis} and detect both of falls while walking and falls while standing\cite{Chaccour2015Smartcarpetusing, Rimminen2010Detectionoffalls, Alwan2006Asmartand, Shin2011Detectionofabnormalliving}. Some studies utilized infrared sensors to detect human movement. M. Popescu \textit{et al.} \cite{Popescu2012VAMPIR-anautomaticfall} classified fall and non-fall using hidden Markov models from infrared sensor signals. A. Yazar \textit{et al.} \cite{Yazar2013Falldetectionusing} utilized support vector machine to detect anomalous signals of vibration sensors while falling after detecting human existence by infrared sensors. S. Tao, M. Kudo and H. Nonaka \cite{Tao2012Privacy-preservedbehavioranalysis} utilized randomized power martingale to discover anomalous falling patterns from discretized values of infrared sensors. J. H. Shin, B. Lee and K. Park \cite{Shin2011Detectionofabnormalliving} utilized infrared motion sensors for each room and detect nine anomalies including falls by support vector data description. Another studies utilized floor sensors or pressure sensors. M. Alwan \textit{et al.} \cite{Alwan2006Asmartand} utilized pattern matching on floor vibrations. H. Rimminen \textit{et al.} \cite{Rimminen2010Detectionoffalls} utilized Markov chain to classify falls and non-falls from floor sensors of electric near field. K. Chaccour \textit{et al.} \cite{Chaccour2015Smartcarpetusing} analyzed the threshold of pressure sensors to detect falls. 

Indoor wandering is also a common anomaly especially among dementia patients \cite{Lin2014Managingelders_wandering}. Many detection methods are proposed \cite{Oliveira2022CNNforElderly, Lin2018Detectingdementia-relatedwandering, Ota2011Elderly-caremotionsensor, Zhao2014Alight-weightsystem, Gochoo2017Device-freenon-privacyinvasive, Chaudhary2020Sensorsignals-basedearly, Khodabandehloo2020Collaborativetrajectorymining} to detect non-routine walking patterns, such as pacing (back-and-forth movement between a start and a goal point), wrapping (moving to a goal point via three or more intermediate locations), and random transitions described by the Martino-Saltzman model \cite{Martino-Saltzman1991TravelBehaviorof}. K. Ota \textit{et al.} \cite{Ota2011Elderly-caremotionsensor} estimated resident movements by measuring the distance from the antenna of an ultra-wideband impulse radio monitoring sensor to classify elderly activities such as sleeping in bed and falling down. W. Zhao \textit{et al.} \cite{Lin2018Detectingdementia-relatedwandering} used active infrared sensors to capture going in and out patterns near the door and classifying walking as wandering when the sensor activation interval is less than 2 minutes and the same sensor reactivated. Q. Lin, W. Zhao and W. Wang \cite{Zhao2014Alight-weightsystem} classified walking patterns using a rule-based method based on the estimated length and frequency of walking trajectories. Many studies \cite{Gochoo2017Device-freenon-privacyinvasive, Chaudhary2020Sensorsignals-basedearly, Oliveira2022CNNforElderly} also utilize neural network models based on the estimated resident's position using infrared motion sensors or simulated ultrasonic sensors. A. Chaudhary \textit{et al.} \cite{Chaudhary2020Sensorsignals-basedearly} employed recurrent neural networks to learn from sensor time series data. M. Goccho \textit{et al.} \cite{Gochoo2017Device-freenon-privacyinvasive} and R. Oliveira \textit{et al.}\cite{Oliveira2022CNNforElderly} employed convolutional neural networks by treating indoor trajectories as images. E. Khodabandehloo and D. Riboni \cite{Khodabandehloo2020Collaborativetrajectorymining} estimated wandering to classify dementia degrees. It used a random forest to classify walk patterns into normal, abnormal or partially abnormal categories based on loop location, center coordinates, and duration of loop-like trajectories that the start and end points of walking matched. Another random forest classified resident's status into healthy, mild dementia or dementia.

Memory impairments can cause interruptions in daily activities for elderly individuals \cite{Yared2016Ambienttechnologyto} including cooking. To prevent fires in the home by forgetting to turn off the oven, sensor systems combining the gas detection and fire detection have been proposed \cite{Logeshwaran2022DesigninganIoT, Hsu2019ApplicationofInternet, Wai2011Pervasiveintelligencesystem, Doyle2014Anintegratedhome-based}. M. Logeshwaran and J. Sheela \cite{Logeshwaran2022DesigninganIoT} proposed an alarm system combining flame sensors, gas sensors, temperature sensors, and PIR sensors. W. Hsu \textit{et al.} \cite{Hsu2019ApplicationofInternet} also proposed similar system with various notification methods including buzzers and LEDs. A. Wa \textit{et al.} \cite{Wai2011Pervasiveintelligencesystem} proposed a system to monitor kitchen usage and heating time. Aside from fire, J. Doyle \textit{et al.} \cite{Doyle2014Anintegratedhome-based} discussed the use of window sensors, door sensors, and power sensors to detect open windows or doors and appliances left on, with a terminal to confirm with residents whether these statuses were intentional. However, quantitative evaluation of these methods are insufficient to detect forgetting in long-term scenario.

While there is no existing research on detecting being semi-bedridden and housebound, there are related methods with these anomalies. J. Williams and D. Cook \cite{Williams2017Forecastingbehaviorin} modeled wake and sleep patterns by predicting each univariate score. T. Hayes \textit{et al.} \cite{Hayes2010EstimationofRest_activity} modeled sleep by four parameters such as start time, end time, sleep latency and total duration time. Some studies detect sleep anomalies after modeling sleep. Z. Shahid, S. Saguna and C. {\AA}hlund \cite{Shahid2022Recognizinglong-termsleep} modeled sleep patterns using three features: start time, duration, and frequency, and detect years-wise sleep behavior changes using clustering analysis. A. Forkan \textit{et al.} \cite{Forkan2015Acontext-awareapproach} modeled the start time of sleep as a normal distribution and considered deviations beyond $\delta_1\sigma$ as a warning as outliers in daily activity statistics. For going out modeling and prediction, S. Tominaga \textit{et al.} \cite{Tominaga2012Aunifiedframework} proposed a generative model from positional trajectory collected by laser range finder.

Beyond typical anomalies, mobility, cognitive function, and mental health levels are also evaluated as predictors for existing assessment scores. A. Akl, B. Taati and A. Mihailidis \cite{Akl2015autonomousunobtrusivedetection} utilized features of walking speed obtained from motion sensor data to perform binary classification of whether an individual has mild cognitive impairment or not, achieving an AUC of 0.97. T. Hayes  \textit{et al.} \cite{Hayes2008Unobtrusiveassessmentof} collected long-term data from groups of mild cognitive impairment (MCI) patients and healthy individuals, investigating the relationship with walking and indoor activities. It concluded that MCI patients tend to have greater fluctuations in median walking speed and more variation in daily activities. C. Galambos \textit{et al.} \cite{Galambos2013Managementofdementia} visually analyze activity levels in the home and going out patterns to evaluate cognitive functions and psychological state. A. Alberdi \textit{et al.} \cite{Alberdi2018SmartHome-BasedPrediction} evaluated motor abilities using the timed up and go (TUG) test and the arm curl test. Cognitive functions are evaluated using the repeatable battery for the assessment of neuropsychological status (RBANS) test, the prospective and retrospective memory questionnaire (PRMQ) test, and the digit cancellation test. Emotional state is measured using the geriatric depression scale 15 (GDS-15) to assess the degree of clinical depression. These values are inferred from behavioral data and used to detect change points of the values. These studies are examples of long-term prediction methods on a weekly basis \cite{Akl2015autonomousunobtrusivedetection, Hayes2008Unobtrusiveassessmentof} and monthly basis \cite{Alberdi2018SmartHome-BasedPrediction}.

For detecting anomalies of various durations, T. Mori \textit{et al.} \cite{Mori2008Anomalydetectionalgorithm} modeled a Gaussian Mixture Model for the start time and duration of estimated activities. This method is used to detect anomalies ranging from several minutes to several days, as well as to detect behavioral variations spanning several months from shifts in distribution. However, they analyze anomalies after detection because they did not aim to detect meaningful typical anomalies but only outliers and shifts. 

From the viewpoint of using anomaly simulations, some studies utilized the predefined hypothesis of anomalies when actual data is difficult to obtain. For wandering detection, two studies \cite{Gochoo2017Device-freenon-privacyinvasive, Chaudhary2020Sensorsignals-basedearly} assigned labels based on decision tree models of movement trajectories \cite{Vuong2014Automateddetectionof}. For fall detection, many studies \cite{Popescu2012VAMPIR-anautomaticfall, Rimminen2010Detectionoffalls, Tao2012Privacy-preservedbehavioranalysis, Yazar2013Falldetectionusing, Chaccour2015Smartcarpetusing} collected fall data by simulating falls through the subject's acting. Other studies \cite{Shin2011Detectionofabnormalliving, Oliveira2022CNNforElderly} used computer simulators to get data. In this context, the simulation has advantages to freely adjust the floor plan, sensors, and abnormal behavior patterns. Similarly, an IoT-based system for early epidemic detection used a discrete event simulator to evaluate abnormal events within activities \cite{Zgheib2023AScalableSemantic}. In our simulation framework \cite{Tanaka2024SensorDataSimulation}, the spatio-temporal variations in anomaly occurrences are broad because anomalies, including a few instances of being housebound and up to a thousand instances of wandering, can occur throughout the home during continuous sensor data over nine years.

\begin{table*}
\centering
\small
\caption{Related works of anomaly behavior detections by ambient sensors (1/2).}
\label{tab:related_works_1}
\begin{tabular}[tbp]{p{1.2cm}p{2cm}p{3cm}p{5.5cm}p{4cm}}
\hline
Anomaly & Reference & Sensors & Methods & Performance\\ \hline
Fall while walking & M. Popescu \textit{et al.}, 2012 \cite{Popescu2012VAMPIR-anautomaticfall} & Vertical PIR sensors array (8 units, 1000 Hz) & Classified higher likelihood using two hidden Markov models with Gaussian output distributions for fall and non-fall & Area under the ROC curve 0.93 for 2-second data, 15\% precision for 26-minute data\\
    & S. Tao, M. Kudo and H. Nonaka, 2012 \cite{Tao2012Privacy-preservedbehavioranalysis} & Horizontal PIR sensors array (20 units, 20 Hz) & Treated 20 sensors as pixel values and used randomized power martingale on strangeness, which is the Euclidean distance from the past average & Precision 92.5\%, recall 98.0\%, F-score 95.14\%\\
    & A. Yazar \textit{et al.}, 2013 \cite{Yazar2013Falldetectionusing} & PIR sensors (256-level output, 100 Hz) and vibration sensors (one per room, range 25 m, 256-level output, 900 Hz) & After detecting human presence when variance of infrared sensor signal exceeds a threshold, it uses Mahalanobis distance, SVM classifies, and single-tree complex wavelet transform & Correctly identified all 60 falls out of 994 data points\\
\hline
Fall while walking and fall while standing & M. Alwan \textit{et al.}, 2006 \cite{Alwan2006Asmartand} & Floor vibration using piezoelectric sensor & Vibration pattern matching using frequency, amplitude, duration, and succession & Correctly detected all 70 falls and 53 object drops. Fisher's exact test shows a true recall range between 94.87\% and 100\%, and true specificity range between 93.28\% and 100\% with a 95\% confidence interval ($p<0.0001$).\\
        & H. Rimminen \textit{et al.}, 2010 \cite{Rimminen2010Detectionoffalls} & Floor sensor of electric near field (9×16 units/$19m^2$, 4.5 Hz) & Two-state (fall, non-fall) Markov chain using number of activated sensors, maximum distance, and total pressure as features & Sensitivity 91\%, specificity 91\%\\
        & J. H. Shin, B. Lee and K. Park, 2014 \cite{Shin2011Detectionofabnormalliving} & 5 PIR sensors (10Hz) for 5 rooms and simulated sensor data  & Support vector data description with the features such as activity level, mobility level and nonresponse duration. & The average performance for the nine anomalies including falls, is a sensitivity of 86\%, a precision of 95.8\% and a specificity of 85.5\%. \\
        & K. Chaccour \textit{et al.}, 2015 \cite{Chaccour2015Smartcarpetusing} & Differential piezoresistive pressure sensors (4 units / 1$m^2$, 21 kHz) & If the values of three or more pressure sensors exceed the threshold, it is considered a fall. & Sensitivity 88.8\%, specificity 94.9\%\\
\hline
Indoor wandering  & K. Ota \textit{et al.}, 2011 \cite{Ota2011Elderly-caremotionsensor} & Ultra-wideband impulse-radio sensor & Estimated actions by linking distance from the antenna with behavior & Classification accuracy including wandering is 95\%\\
    & W. Zhao \textit{et al.}, 2014 \cite{Zhao2014Alight-weightsystem} & PIR sensors & Rule-based classification using trajectory position and duration time of stay & Accuracy 90.03\%\\
    & M. Gochoo \textit{et al.}, 2017, \cite{Gochoo2017Device-freenon-privacyinvasive} & PIR sensors & Convolutional neural network using binary images converted from binary sensor activation sequences & Accuracy 97.84\%, precision 97.9\%, recall 97.8\%\\
    & Q. Lin, W. Zhao, and W. Wang, 2018 \cite{Lin2018Detectingdementia-relatedwandering} & Active infrared sensors & Classified as wandering when sensor activation interval is less than 2 minutes and the same sensor reactivates & Accuracy 98\%, precision 98\%\\
    & A. Chaudhary \textit{et al.}, 2020,\cite{Chaudhary2020Sensorsignals-basedearly} & PIR sensors & Recurrent neural network Learned by binary sensor activation sequences & Accuracy 98.63\%\\
    & R. Oliveira \textit{et al.}, 2022 \cite{Oliveira2022CNNforElderly} & Simulated ultrasonic sensors for human position and microphone recording for 5 seconds after proximity detection & Convolutional neural network using images of hourly movements & Precision 100\%, recall 60\%, F-score 75\%\\  
\hline
Indoor wandering and dementia & E. Khodabandehloo and D. Riboni, 2020 \cite{Khodabandehloo2020Collaborativetrajectorymining} & PIR sensors (51 units) and door sensors (16 units) in a two-story house & Firstly, it estimated loop-like trajectories from sensor data where start and end points of walking match. Then, it classified walking into three categories (normal, abnormal, partially abnormal) using random forest based on features of loop location, center coordinates, and duration. It also classified dementia status (healthy, dementia, mild dementia) from walking classifications using random forest. & Average recall for wandering is 60.5\%, average precision is 60.2\%. Classification accuracy for dementia is 80.7\%.\\
\hline
\end{tabular}
\\
\flushleft{Abbreviations are passive infrared (PIR).}
\end{table*}

\begin{table*}
\centering
\small
\caption{Related works of anomaly behavior detections by ambient sensors (2/2).}
\label{tab:related_works_2}
\begin{tabular}[tbp]{p{2.4cm}p{1.8cm}p{3cm}p{4.9cm}p{3.9cm}}
\hline
Anomaly & Reference & Sensors & Methods & Performance\\ \hline
Forgetting to turn off & A. Wai \textit{et al.}, 2011 \cite{Wai2011Pervasiveintelligencesystem} & Motion sensor, current sensor, temperature 
sensor & Detects absence of people with motion sensor, alerts with power and high temperature & No quantitative evaluation \\
    & J. Doyle \textit{et al.}, 2014 \cite{Doyle2014Anintegratedhome-based} & Window sensor, door sensor, power sensor & Notifies residents of open windows or lights left on through devices & No quantitative evaluation \\
    & W. Hsu \textit{et al.}, 2019, \cite{Hsu2019ApplicationofInternet} & Flame sensor, gas sensor, temperature sensor & Alerts when flame, gas leak, or high temperature is detected by sensors & No quantitative evaluation\\
    & M. Logeshwaran and J. Sheela, 2022 \cite{Logeshwaran2022DesigninganIoT} & Flame sensor, gas sensor, temperature sensor, PIR sensor & Alerts when flame, gas leak, or high temperature is detected by sensors & No quantitative evaluation\\
\hline
Cognitive function & T. Hayes \textit{et al.}, 2008 \cite{Hayes2008Unobtrusiveassessmentof} & PIR sensor (1/6Hz), magnetic contact sensor & Estimates features related to walking speed and activity level and analyzes the correlation with mild cognitive impairment & For mild cognitive impairment group, the coefficient of variation of median walking speed is twice as high compared to healthy group ($p<0.03$), 24-hour wavelet variance is larger (greater variation in daily activity patterns) ($p<0.008$)\\
                   & A. Akl, B. Taati, and A. Mihailidis, 2015, \cite{Akl2015autonomousunobtrusivedetection} & PIR sensor & Estimates features related to walking speed and activity level and trains a binary classifier (for healthy and mild cognitive impairment) using SVM and random forest & For binary classification of mild cognitive impairment, area under ROC curve is 0.97, area under precision-recall curve is 0.93 \\
\hline
Cognitive function and psychological state & C. Galambos \textit{et al.}, 2013 \cite{Galambos2013Managementofdementia} & PIR sensor (1/7Hz) & Visualizes activity level by showing values proportional to the number of sensor activations in each room on a 2D coordinate system of day and time, visually checks for abnormalities & Qualitatively evaluates the visualization of activity level and going out, comparing with unhealthiness in each case \\
\hline
Motor and cognitive functions, psychological state & A. Alberdi \textit{et al.}, 2018 \cite{Alberdi2018SmartHome-BasedPrediction} & PIR sensor, door (cupboard) sensor, e.g., 23 PIR sensors and 6 door sensors in a layout with bedroom, kitchen, dining room, living room, and bathroom & Firstly, it assigns behavior labels using activity recognition method \cite{Krishnan2014Activityrecognitionon}. Then, it regresses test item values from features of behavior, movement, and habitual behavior using kNN, and detects score differences with the previous day using Reliable Change Index \cite{Christensen1986Amethodof} & Statistically significant detection of change points in Arm Curl, TUG, and delayed memory impairment items in RBANS, under p-value adjusted by McNemar test compared to random algorithm ($<0.005$)\\
\hline
Being semi-bedridden, being housebound, forgetting to turn off, wandering, fall while walking and fall while standing & Proposed method & Simulated sensor data of PIR sensor (10Hz), door sensor (10Hz), pressure sensor (10Hz), power sensor (1Hz), and water flow sensor (1Hz) & Statistical test methods for being semi-bedridden and being housebound, decision tree for forgetting, random forest for fall while walking, hidden Markov model for fall while standing and wandering. & (Sensitivity and false alarm rate per day) are (1.0, 0) for being semi-bedridden, (1.0, 0.004) for being housebound, (1.0, 0.01) for forgetting, (1.0, 0.017) for wandering, (0.75, 0.02) for fall while walking, and (0.92, 0) for fall while standing.\\
\hline
\end{tabular}
\\
\flushleft{Abbreviations are; clinical dementia rating (CDR), geriatric depression scale (GDS), mini-mental state examination (MMSE), passive infrared (PIR), prospective and retrospective memory questionnaire (PRMQ), repeatable battery for the assessment of neuropsychological status (RBANS), short-form health survey (SF-12), and timed up and go (TUG).}
\end{table*}

\section{System overview}
We propose a simulator-based anomaly detection system that addresses the issue of data insufficiency in smart homes by utilizing predefined anomaly models and sensor data simulator. An overview of this system is shown in Fig. \ref{fig:system_overview}. First, the simulator is customized by setting parameters to values appropriate for the given operational environment: [layout \& sensors] room size and placement of furniture, sensors' models, numbers and arrangement, [resident's information] activity statistics, including start time, duration and frequency, and walking speed, [model of anomalies] degree of dementia, types of anomalies, observed phenomena, and frequency. Second, time series data of sensor outputs and anomaly labels are generated by the customized simulator based on resident behavior models and anomaly models. Third, the generated time series data are transformed into feature vector sequences and label interval sequences appropriate for each anomaly. The detection classifier for each anomaly is learned using the transformed training feature vector and label interval data in the final stage of learning phase. In operation, real sensor time series data are input and transformed into feature vector data for each anomaly, and the label of each anomaly is predicted by the trained classifier. The preprocessing procedure for the anomaly detection method is detailed in Fig. \ref{fig:preprocessing}.

\section{Data simulation}

Data are generated using an existing simulator \cite{Tanaka2024SensorDataSimulation}. This simulator has five components; (1) floor plan and sensor arrangement, (2) activity, (3) walking trajectory, (4) sensor activation, and (5) anomaly modeling. In this experiment, each component is configured as follows.

\subsection{Anomaly modeling}
Anomalies affect simulation processes and are implemented within other simulator components. Some anomalies depend on the latent dementia level, simulated as the MMSE (referred to as $M$ in Tab. \ref{tab:six_implemented_anomalies}) determined by a random autoregressive model \cite{Tanaka2024SensorDataSimulation}. The implemented models are summarized in Tab. \ref{tab:six_implemented_anomalies}.

\subsection{Floor plan and sensor arrangement}
An example of the floor plan and sensor arrangement is shown in Fig. \ref{fig:layout_and_sensor_arrangement}. We assume the resident lives in a studio apartment without any partitions between rooms. As ambient sensors, we use sensors, including infrared motion sensors, pressure sensors, door sensors, power sensors, and flow sensors as listed in Tab. \ref{tab:list_of_ambient_sensors}. The sensors are arranged manually to cover the main flow line in the layout.

\subsection{Activity generation}
The activity generator is parameterized by the statistics of the activities of daily living (ADL). For example, Fundamental activities, such as sleeping and eating, can be defined by their start time and duration of the activity using normal distributions. Each activity is assigned daily to guarantee its statistics. These activities are considered as presence patterns within specific locations, such as eating at the dining table.

Being semi-bedridden and being housebound are assumed to be anomalies that affect daily activity statistics over several weeks. The symptoms of being semi-bedridden, which occur once every 20 months and continues 30 days on average, are assumed that a 40 minute nap is added, rest time increases by 30 minutes compared to normal, and outing frequency changes to once a week on average. The symptoms of being housebound, which occur once every 10 month and continues 14 days on average, are assumed that the frequency of social activities decrease such as using the phone occurs on average once every three days, and going out occurs once every two weeks.

\subsection{Walking trajectory}
According to the activity schedule, the walking patterns where the resident moves between locations when switching to the next activity are simulated. In this simulation, the resident is assumed to walk from the previous activity location to the next activity location at a fixed walking speed, and the next activity is assumed to begin when the resident reaches within a 30 cm range of the destination.

Wandering, fall while walking, and fall while standing are assumed to be related to walking patterns. Wandering is implemented as aimless random movement among various activity locations within the room. The frequency and duration of wandering increase proportionally with MMSE $M$, with an average of $-1.86M + 56$ occurrences per month, and an average duration of $-0.31M + 9.8$ minutes per occurrence. In this study, falls are assumed to represent the total time from the moment of falling until getting up and restarting to walk. Falls while walking occur at a random location in a walking trajectory, while falls while standing occur at the starting location of a walk. Both types of falls are assumed to be stopped for an average of 30 seconds at the falling location. The frequency of falls increases proportionally with MMSE, with an average of $-M/15 + 2$ occurrences per month for each.

\subsection{Sensor activation}
All sensors have binary states (on and off). Infrared motion sensors are activated when the resident moves within the detection range and deactivate when the movement stops. Pressure sensors are activated while the resident is standing on them. Door sensors are activated when the resident uses the entrance to go in or out. Power sensors and flow sensors activate while the resident uses home appliances or faucets. These sensor data are recorded as a sequence of tuples containing the activation time, sensor index, and sensor state, e.g., (10days 13:45:30.1, 10, ON).

Forgetting to turn off the home appliances and faucets affects the sensor activation patterns. It occurs an average of $-M+30$ times per month. When the forgetting occurs, one activity in which a home appliance is used is randomly selected, then the home appliance is forgotten to turn off until the resident comes back to near the area.

\begin{table*}
\centering
\caption{List of ambient sensors with binary outputs used in this study.}
\label{tab:list_of_ambient_sensors}
\begin{tabular}[tbp]{p{3cm}p{1.2cm}p{5.5cm}p{6.5cm}}
\hline
Sensor& Sampling rate & Place & Detected target\\ \hline
Infrared motion sensor & 10Hz & Open area with a detection range of 50 cm radius & Resident motion\\
Pressure sensor & 10Hz & Floor near the bed with a detection range of 0.75$\textrm{m}^2$ and 1$\textrm{m}^2$ & Resident presence\\
Door sensor & 10Hz & Entrance door & Door openings and closings\\
Flow sensor & 1Hz & Faucets in bathroom and kitchen & Captures water usage, similar to a utility meter\\
Power sensor & 1Hz & TV and kitchen stove & Captures power consumption, similar to a utility meter\\
\hline
\end{tabular}
\end{table*}

\begin{figure}[tbp]
\centering
\includegraphics[width=120mm]{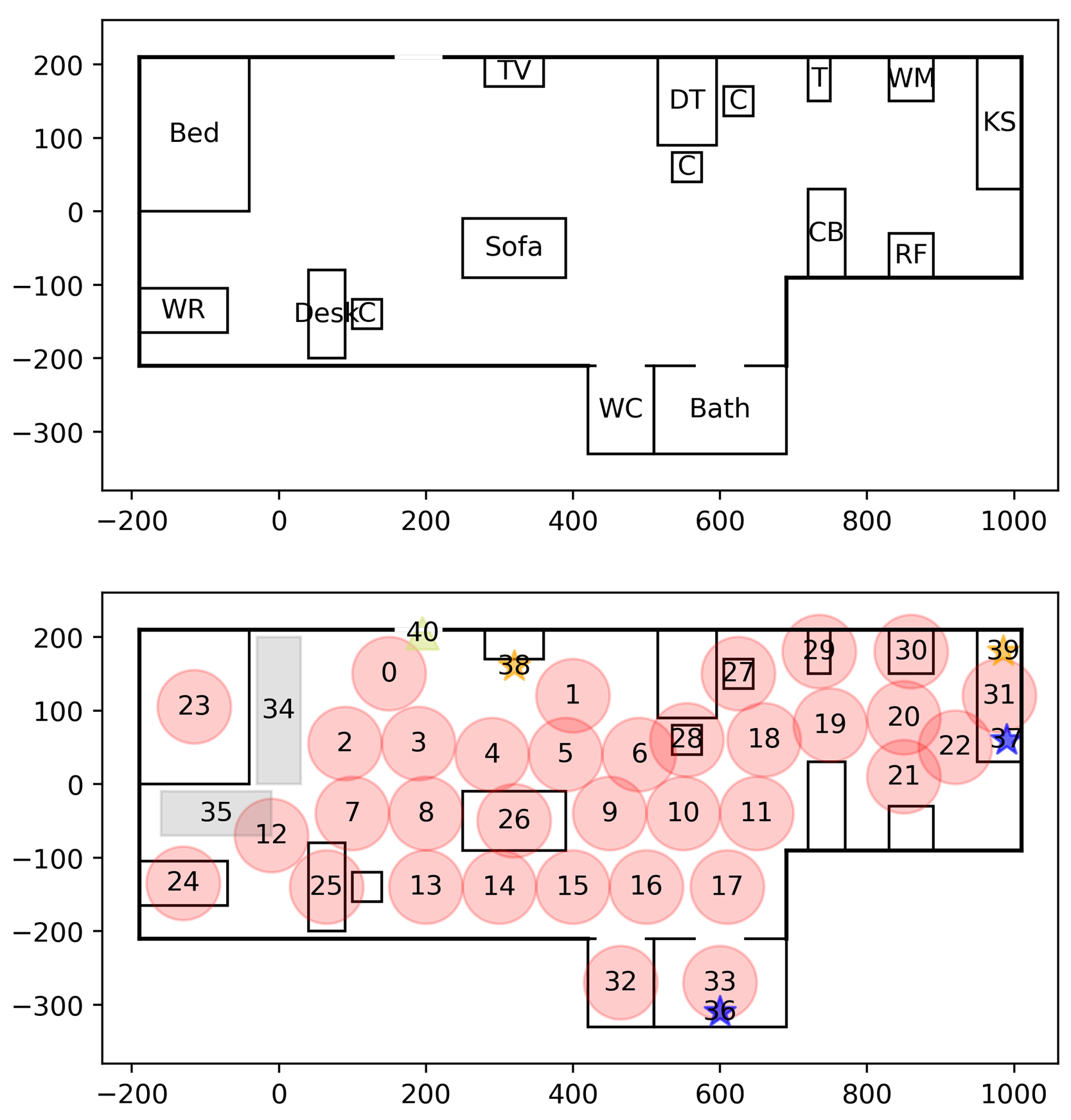}
\caption{Example of a floor plan and sensor arrangement. The layout is 5 meters by 12 meters. There are chairs (C), a cupboard (CB), a dining table (DT), a kitchen stove (KS), a refrigerator (RF), a trash box (T), a wardrobe (WR), a washing machine (WM) and a water closet (WC). Colors and symbols differentiate sensor types: red circles for passive infrared sensors (\#0 to \#33), gray squares for pressure sensors (\#34 and \#35), blue stars for flow sensors (\#36 and \#37), yellow stars for power sensors (\#38 and \#39), and green triangles for the door sensor (\#40)}
\label{fig:layout_and_sensor_arrangement}
\end{figure}

\section{Feature constructions and classifiers}

\subsection{Pre-processing}
\begin{figure*}[tbp]
\centering
\includegraphics[width=\linewidth]{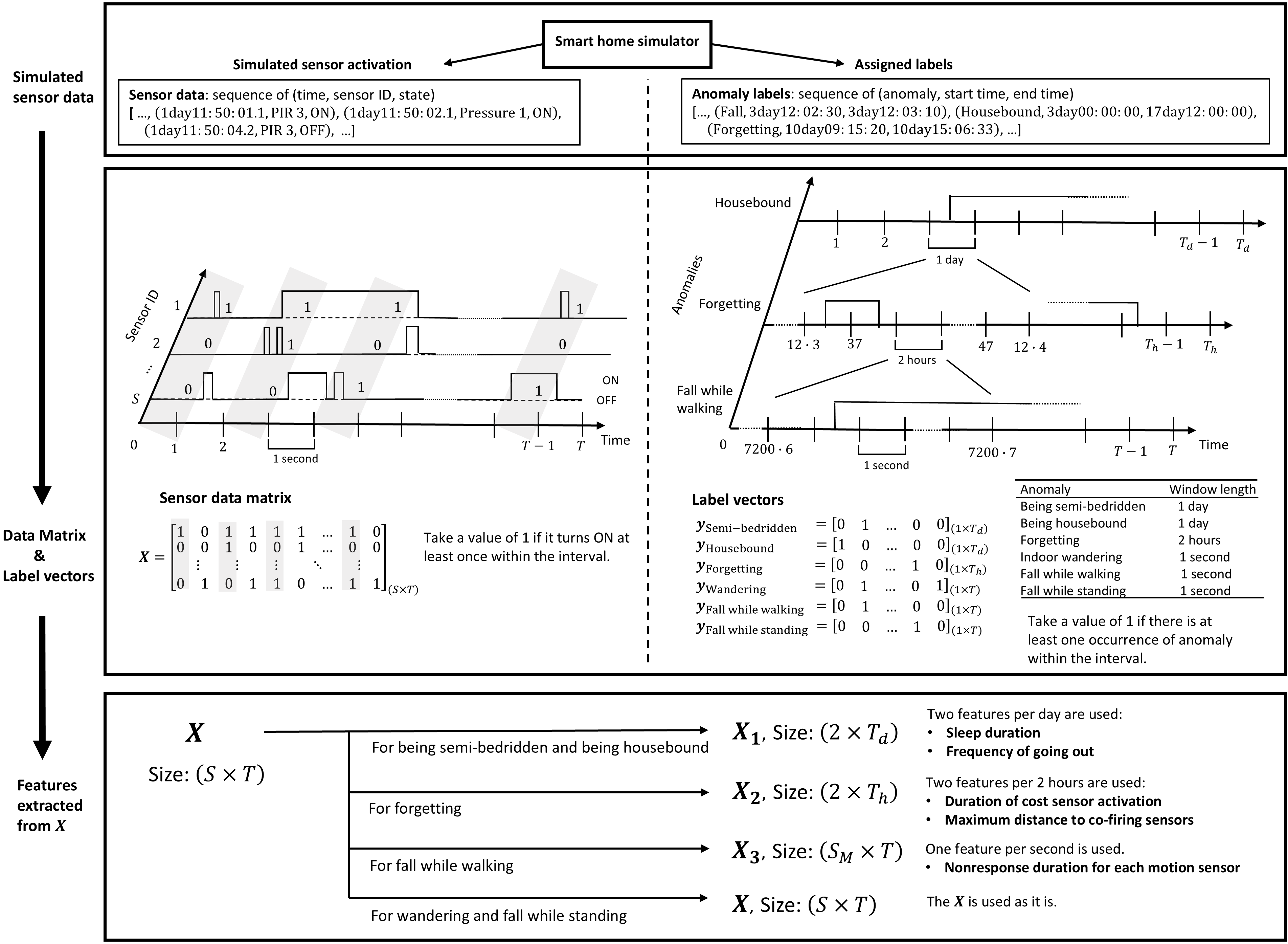}
\caption{Illustration of our training data construction using a simulator. Initially, the smart home simulator generates a time series data of $S$ sensor firings and anomaly labels. The sensor data is summarized into a one-second interval binary-value vector sequence (data matrix). Each anomaly label sequence is summarized into a binary-value sequence with anomaly-specific intervals. To learn a classifier for each of the six anomalies, the sequence of feature vectors appropriate for each anomaly is extracted from the sensor data matrix.}
\label{fig:preprocessing}
\end{figure*}

Sensor data and anomaly labels are simulated as time series data. Sensor data is initially given as a sequence of tuples containing the time, sensor ID, and sensor state (1: ON, 0: OFF) as shown in the left part of the top box of Fig. \ref{fig:preprocessing}. Binary sensors used in this study are shown in Tab. \ref{tab:list_of_ambient_sensors}: infrared motion sensors, pressure sensors, door sensors, flow sensors and power sensors. In the simulation, six abnormal behaviors (as listed in Tab. \ref{tab:six_implemented_anomalies}) are generated according to our anomaly models. Due to the simulation, true labels of anomalies are also acquired as a sequence of tuples containing the time and status indicating whether anomalies occur (1: occurred, 0: not occurred) as illustrated in the right part of the top box of Fig. \ref{fig:preprocessing}).

Raw data is preprocessed to obtain matrix forms. Sensor data is summarized every second. Let $T$ be the total number of seconds in the data, then, the sensor data are represented by a matrix $\oMat{X}=(\oMat{X}_{ij}) \in \{0, 1\}^{S \times T}$, where $i$ denotes the $i$th sensor $(i=1, 2, \ldots, S)$, and $j$ denotes the $j$th second $(j=1, 2, \ldots, T)$. Here, $\oMat{X}_{ij}=1$ indicates that the $i$th sensor was activated at least once during the interval from the $(j-1)$th to the $j$th second (e.g., the left part of the middle box of Fig. \ref{fig:preprocessing}); otherwise, it is 0. This data representation was used in some studies \cite{Van2011Humanactivityrecognition}. Similarly, label data for each anomaly is also summarized. However, the interval length varies depending on the anomaly (one second for falls and wandering, two hours for forgetting, and one day for housebound and semi-bedridden as shown in the right part of the middle box of Fig. \ref{fig:preprocessing}). As a result, the lengths of the label vectors $\oVec{y} \in \{0, 1\}^{L}$ are also different ($L$ is equal to $T$ for falls and wandering, $T_h(=T/7200)$ for forgetting, and $T_d(=T_h/12)$ for being housebound and semi-bedridden).

\subsection{Detection of falls while waking}

To detect falls while walking, we use random forests \cite{Breiman2001RandomForests} with nonresponse duration as a feature to focus on detecting resident immobility. We extract the feature from all $S_M$ motion sensors including infrared motion sensors, pressure sensors, and door sensors. The nonresponse duration for the $i$th motion sensor is the time elapsed since the $i$th sensor was last activated among motion sensors until any other motion sensor is activated. This duration can be interpreted as an approximation of the elapsed time the residents did not move. We changed the definition slightly according to the motion sensor types. Because infrared sensors and door sensors capture instantaneous movements of a person, the elapsed time is reset when the same sensor is activated again. In contrast, pressure sensors detect continuous presence, so the elapsed time is not reset when the same sensor is activated again.

We then construct features as $\oMat{NRD}\in \NaturalNumber^{S_M\times T}$ from $\oMat{X}$, where $\oMat{NRD}_{i, j}$ represents nonreponse duration of sensor $i$ at $j$th seconds. The random forest $f$ learns the binary classification $\function{f}{\NaturalNumber^{S_M}}{\{0, 1\}}$ by treating each second as a single instance.

\subsection{Detection of falls while standing and wanderings}

To detect falls while standing and wandering, we use the data matrix and label vector without additional preprocessing. These data are modeled by standard time series classifiers \cite{Van2011Humanactivityrecognition} such as dynamic naive Bayes \cite{Aviles2011Acomparisonof} and hidden Markov model \cite{Rabiner1989Atutorialon}. To suppress noisy predictions, continuous labels with durations shorter than the given denoising threshold are removed as a post-processing step.

\subsection{Detection of forgetting to turn off the home appliances and faucets}

To detect instances of forgetting to turn off appliances and faucets, we use the following two types of features calculated every 2-hour interval.
\begin{itemize}
\item \textbf{Duration of cost sensor activation}: The sum of the time period for which the appliance sensor is activate.
\item \textbf{Maximum distance to sensors activated during cost sensor activation}: The maximum Euclidian distance from the cost sensor to any other sensor that was also activated during the time the cost sensor was activated.
\end{itemize}

Decision tree \cite{Loh2011ClassificationAndRegression} is learned with the two-dimensional feature vector to classify the binary label. The splitting criterion is Gini impurity, and the stopping criteria requires each leaf node to contain less than five data points.

\subsection{Detection of being semi-bedridden and being housebound}

We propose a method that determines being semi-bedridden or being housebound based on a statistical test approach. Both being semi-bedridden and housebound involve a reduction in the frequency of going out. To distinguish the semi-bedridden state from the housebound state, we first classify the semi-bedridden state using sleep duration time and then classify periods with low frequencies of going out as housebound state. These classification interval is set to one day.

We explain the classification method step by step below.
\begin{enumerate}
\item \textbf{Rule-based estimation of daily sleep duration} If motion sensors near the bed (\#23, \#34, and \#35 in the case of Fig. \ref{fig:layout_and_sensor_arrangement}) are activated, and the interval until these sensors are next activated exceeds one minute, this interval is considered as the sleeping time. However, if any other motion sensors outside of the bed are activated during this interval, it is not considered as sleep.
\item \textbf{Modeling the sleep duration} The daily sleep duration is represented by a normal distribution with the mean $\mu_s$ and standard deviation $\sigma_s$, where $\mu_s$ and $\sigma_s$ are estimated by the sample mean and sample standard deviation of all the estimated daily sleep durations.
\item \textbf{Learning the semi-bedridden classification} Periods where the sleep duration exceeds the threshold $\theta_s = \mu_s + c_s \sigma_s$ for more than seven consecutive days are classified as semi-bedridden. The parameter $c_s$ is selected from the closed interval $[-1, 3]$ to maximize the F1 score in the training data.
\item \textbf{Rule-based estimation of daily frequency of going out} If the door sensor is activated and the interval until the sensor is next activated exceeds one minute, it is counted as one instance of going out. However, if any other motion sensors are activated during this interval, it is not counted.
\item \textbf{Modeling the going out} The daily frequency of going out is modeled by a normal distribution with the mean $\mu_h$ and standard deviation $\sigma_h$, where $\mu_h$ and $\sigma_h$ are estimated by the sample mean and sample standard deviation of all the estimated daily going-out frequencies.
\item \textbf{Learning the housebound classification} Periods where the frequency of going out falls below the threshold $\theta_h = \mu_h - c_h \sigma_h$ for more than seven days and are not classified as semi-bedridden are classified as being housebound. The parameter $c_h$ is selected from the closed interval $[-1, 3]$ to maximize the F1 score in the training data.
\end{enumerate}

Note that we now hypothesize that the bed is used exclusively for sleep, not for other activities such as reading.

\section{Experiments}
To demonstrate the effectiveness of our proposed simulator-based anomaly detection system, we evaluate its anomaly detection performance for a typical setting for room layout, sensor arrangement, and an elderly resident, using simulator-generated test data. 

\subsection{Simulation data}
We prepare the simulation data under the following settings. First, floor plan and sensor arrangement are predefined as shown in Fig. \ref{fig:layout_and_sensor_arrangement}. We placed 41 sensors to cover the main flow lines in the floor plan. In the house, the resident follows an activity model with default parameters used in the study \cite{Tanaka2024SensorDataSimulation} and proposed anomaly models. As a latent value for anomalies, the MMSE reduces from 29 to 19.5 over nine years on average, reflecting the progression rate at the early stage of dementia \cite{Barry2010StagingDementia}. In this model, the resident moves at a fixed walking speed of 68.75 cm/s. Sensors are simulated at 10 Hz for infrared/pressure sensors and 1 Hz for door/cost sensors. We simulated 9 years ($9\cdot 360$ days) both for training and test data. The training dataset contains 7555956 sensor activations. The numbers of anomaly labels in training data and test data are; 4 and 8 for being semi-bedridden, 11 and 17 for being housebound, 452 and 575 for forgetting, 952 and 955 for wandering, 26 and 40 for falls while walking, and 32 and 38 for falls while standing. The simulator code and data is in public at \textrm{https://github.com/tanakai0/SensorDataSimulator}.

\subsection{Classifiers}
For detecting being semi-bedridden and housebound, the accuracies of estimated daily sleep duration and going-out frequency are calculated by the mean absolute errors (MAE) between the estimated values from the simulated sensor data and the true values from the activity labels generated by the resident behavior model in the simulation. The MAE values were 6.8 minutes and 0.046 occurrences in the training data, and 6.9 minutes and 0.042 occurrences in the test data. The model for estimated sleep duration and going-out frequency are $N(8.02, 1.15)$ and $N(3.95, 2.12)$, respectively, and learned parameters are $c_s=0.10$ and $c_h=1.80$. As a result, the thresholds are calculated as $\theta_s=8.13$ and $\theta_h=0.14$.

For detecting falls while walking, a random forest of 100 decision trees is trained. To reduce the feature vectors corresponding to the 0 anomaly labels, the random forest is trained using data from the 3 days before and after each fall occurrence time.

\subsection{Evaluation metrics}

Let $\oVec{y}$ and $\hat{\oVec{y}}$ represent the sequences of true (model-generated) and predicted anomaly labels. We use five evaluation metrics for alarms of anomaly behaviors. The two metrics are typical precision and recall between $\oVec{y}$ and $\hat{\oVec{y}}$. 

The other three metrics are defined using the \textit{label interval} sequence of $\oVec{y}$,
which is defined as $I=\{I_1, I_2, \ldots, I_N\}$, where $I_i$ is the $i$th maximal time interval
of consecutive 1s in $\oVec{y}$. For example, the label interval sequence of $\oVec{y} = [0, 1, 1, 1, 0, 0, 1, 1, 0, \ldots]$ consists of $I_1 = [1, 3], I_2 = [6, 7], \ldots$. Let $I=\{I_1, I_2, \ldots, I_N\}$ and $\hat{I} = \{\hat{I}_1, \hat{I}_2, \ldots, \hat{I}_{\hat{N}}\}$ be the label interval sequences of $\oVec{y}$ and $\hat{\oVec{y}}$, respectively. Note that $I$ and $\hat{I}$ are composed of disjoint intervals.

\begin{itemize}
  \item \textbf{Sensitivity} is the proportion of correctly detected anomalies, which is defined as \\$\lvert\{I_{a}\mid I_{a}\cap \hat{I}_{b}\neq \emptyset, \ForSome {b}\}\rvert$ / N. This measure is an interval version of recall, and high sensitivity indicates that a high proportion of anomalies are detected.
  \item \textbf{False alarm rate (FAR)} represents the proportion of wrongly detected anomalies and given as \\
  $\lvert\{\hat{I}_{b}\mid I_{a}\cap \hat{I}_{b} =\emptyset, \ForAny {a}\}\rvert$ / (the number of days),\\
  i.e., it reflects the number of false alarms per day. Low FAR means that there are few unnecessary warnings.
  \item \textbf{Mean alarm length (MAL)} is the average of predicted label intervals, regardless of whether they are correct or not, and defined by $\left(\sum_{{b}=1}^{\hat{N}}\lvert\hat{I}_{b}\rvert\right) / \hat{N}$, where $\lvert\hat{I}_{b}\rvert$ denotes the length of $\hat{I}_{b}=[s, e]$, i.e., $\lvert\hat{I}_{b}\rvert=e-s$. This should be minimized to reduce redundant alarms. Note that extending predicted label intervals makes sensitivity and FAR improve, but MAL deteriorates.
\end{itemize}
When calculating above three metrics for wandering and falls, denoise thresholds are used to remove the noisy labels shorter than the threshold. For example, 5 seconds threshold remove the label intervals shorter than 5 seconds.

A similar evaluation metrics were used in the existing study for long-term fall detection in real world \cite{Kangas2014Sensitivityandfalse}. This study experimented more than three years using wearable sensors for fall detection, and achieved a sensitivity of 80\% with 1 false alarm per 40 usage hours.

\subsection{Result}

The results are shown in Tab. \ref{tab:classification_report}. Note that evaluation metrics are different from those used in existing works in Tab. \ref{tab:related_works_1} and Tab. \ref{tab:related_works_2}, making direct comparisons of methods difficult. As the simple numeric comparison, the performance in detecting falls and wandering is comparable to existing works. The other three newly detected methods for being semi-bedridden, being housebound, and forgetting achieve a sensitivity of over 0.9, with false alarms occurring less than once every 100 days. For detecting wandering and fall while standing, the hidden Markov models outperform the dynamic naive Bayes because it can utilize the time context. Detecting falls while walking remains the most challenging with a sensitivity of 0.7. This is due to the difficulty in distinguishing whether the resident has fallen or merely reached their destination when he or she has stopped walking near activity areas. However, over sampling labels are promising to improve the performance. In this study, the training data included 26 instances over 9 years while previous research \cite{Tanaka2024FallDetectionby} has achieves the sensitivity 0.96 by oversampling the number of falls around 200 times.

\begin{landscape}
\begin{table*}
\centering
\caption{Test results of detections for six types of anomalies from nine years of a simulation data. The evaluation metrics are calculated by the true labels on the simulation data.}
\label{tab:classification_report}
\begin{tabular}[tbp]{p{2.2cm}p{1.5cm}p{1.0cm}p{1.2cm}p{1.3cm}p{1.3cm}p{1.8cm}p{1.1cm}p{1.2cm}p{1.6cm}p{2.0cm}}
\hline
Anomaly & Unit length & \#Anoms. & Methods & Raw precision & Raw sensitivity & Denoise threshold & Precision & Sensitivity & FAR [\#alarms/day] & Mean alarm length \\ \hline
Semi-bedridden      & 1 day    & \hphantom{00}8 & ST & 0.97  & 0.80 & - & 1.0 & 1.0 & 0.0 & 14.9 days\\\hline
Housebound          & 1 day    & \hphantom{0}17 & ST & 0.82  & 1.0  & - & 0.54 & 1.0 & 0.004 & \hphantom{0}9.2 days\\\hline
Forgetting          & 2 hours  &            575 & DT & 0.98  & 0.93 & - & 0.74 & 1.0 & 0.01 & 23.4 hours\\\hline
Wandering           & 1 second &            955 & DNB & 0.10 & 0.53 & - & 0.10 & 1.0 & 68.2 & 2.78 seconds\\
                    &          &                &     &      &      & 5 seconds & 0.10  & 0.97 & 13.1 & 5.97 seconds\\
                    &          &                & HMM & 0.10 & 1.0  & - & 0.006 & 1.0 & 45.9 & 8.85 seconds\\
                    &          &                &     &      &      & 28 seconds & 0.94 & 1.0 & 0.017 & 144 seconds\\\hline
Fall while walking  & 1 second & \hphantom{0}40 & RF & 0.20  & 0.64 & - & 0.092 & 0.75 & 0.09 & 12.2 seconds\\
                    &          &                &    &       &      & 16 seconds & 0.32 & 0.75 & 0.02 & 29.6 seconds\\\hline
Fall while standing & 1 second & \hphantom{0}38 & DNB & 0.83 & 0.20 & - & 0.13 & 0.18 & 0.015 & 5.13 seconds\\
                    &          &                & HMM & 0.66 & 0.92 & - & 0.17 & 0.92 & 0.053 & 7.7~~seconds\\
                    &          &                &     &      &      & 6 seconds & 0.97 & 0.92 & 0.0 & 30~~~~seconds\\\hline
\end{tabular}
\\
\flushleft{Abbreviations are; number of anomalies in test data (\#Anoms.), statistical test (ST), decision tree (DT), random forest (RF), dynamic naive Bayes (DNB), and hidden Markov model (HMM). Five evaluation metrics are raw precision, raw recall, sensitivity, false alarm rate (FAR) and mean alarm length. Raw precision and raw recall are calculated using labels for data segments of unit time length. Precision, sensitivity, FAR and mean alarm length are calculated based on the label intervals, which correspond to consecutive anomaly data segments. For the latter metrics, label intervals shorter than the denoising threshold are treated as noise and excluded.}
\end{table*}
\end{landscape}

\section{Discussion}

\subsection{Use case}
We discuss how this system could be used in a real scenario. There is an elderly individual living alone who is either healthy or early stage of mild cognitive impairment but still capable of living independently. The person, family member, or caregivers who visit regularly, wishes to monitor long-term abnormal behaviors over several years to detect early signs of health deterioration. This system records the anomaly predictions and alert to the person, family member, caregivers, or service manager. As used in \cite{Galambos2013Managementofdementia}, anomaly-driven video sensors may preserve the resident's privacy except in urgent situation. For abnormal behaviors that require immediate attention, such as falls or forgetting to turn off appliances, the system could trigger alerts like a buzzer for the individual or make phone calls to relatives. Even for behaviors that do not require urgent response, such as being housebound or being semi-bedridden, health data recorded by the system can be shared for visits or online tools to access detailed health evaluations without overlooking. This approach helps prevent further deterioration of the individual’s health.

We also discussed how the system's simulators and detection methods can be personalized. When conducting data simulations in a real-world environment, the necessary data includes the floor plan, sensor placement, and the resident’s behavioral statistics. In this study, the behavioral and movement patterns were simulated using average behavioral statistics of elderly individuals, that can be applied to broad target persons. However, if data such as activity statistics (e.g., when the person sleeps or how many times the person goes out) or walking speed can be collected, simulator’s generation model can incorporate those to personalize the system for the resident.

\subsection{Performance gap from evaluation using real data}

The performance of the proposed methods is evaluated using simulation test data. How much differences are there from those calculated using real test data? If the performances change significantly, then our experimental results are not reliable at all and meaningless. Evaluation using real test data must be done in the future, but the long-period experiments using human subjects are ethically difficult and some anomaly occurs rarely, thus evaluation using real data seems unable. Here, we discuss degree of reliability of our experimental results.

Our simulator assumes that the resident takes daily actions based on an activity model, moves on foot following a walking model, performs anomaly actions based on an anomaly model, and is detected by a sensor model. Action sequence generated by our activity model has been already demonstrated to be similar to real activity sequence in \cite{Tanaka2024SensorDataSimulation}. For our binary-valued sensors, we do not need an intricate walking model, and walking speed and direction are enough. We assume that the resident walks to the next activity location through the shortest path at the fixed speed, which seems reasonably close to a real walking trajectory. As for simulated anomaly action sequence, its similarity to real data has not been checked yet, and anomaly models can be improved by checking the similarity, but such real data is difficult to collect now. With respect to our long duration anomalies (semi-bedridden, housebound, forgetting and wandering), action sequences generated by our anomaly models when the anomalies occur, seem not so different from real action sequences. When falling occurs, however, the simulator-generated action sequences might be different from real action sequences to some extent and should be checked in the future. Since we use simple binary sensors that is turned on by the resident's activity and moving, sensor value sequence generated by our sensor model is similar to the real sensor value sequence under the same resident's activity and moving sequence.

Comprehensively considering, we believe that performance results using simulation test data are not so different from those using real test data except for falling detection, whose performance might be also similar to those using real test data but checking the difference and improving the model if necessary should be done in the future.

\subsection{Expandability of the model}

There are several ways to enhance the realism such as enhancement of activity and anomaly models, introduction of data shift in real world, and feedback from real data.

For current activity models, the start time, duration, and frequency of activities and anomalies are assumed to follow simple distributions, such as Poisson or normal distributions. However, more detailed distributions, such as those used in \cite{Tominaga2012Aunifiedframework, Williams2017Forecastingbehaviorin, Hayes2010EstimationofRest_activity}, can also be employed in the perspective of personalization. For falls, we simply define them as periods of immobility after the falling. In reality, residents may attempt to move, such as by crawling, after a fall. Residents tend to fall in a particular locations such as narrow spaces, steps, and areas with scattered objects. Even if there is the argument that simulation cannot fully replicate real-life characteristics of falls \cite{Stack2017Fallsareunintentional}, prior data and knowledge of falls could improve the similarity to the real.

There are several unintended events or shifts in real life, such as sensor failure due to light, heat, or hardware issues; resident irregular activity including sleep on the floor; and seasonal shift including sleep in sofa in the summer. These types of anomalies could potentially be addressed by sensor malfunction detection through federated learning \cite{Sater2021Afederatedlearning} and concept drift detection methods for activity patterns \cite{Friedrich2023UnsupervisedStatisticalConcept}.

Verification and feedback in the real world are promising to enhance the model. Collecting real sensor data labeled with target anomalies would directly improve the anomaly model, but this process is costly. To reduce costs, it is promising to use partial labeling such as data derived from diary reports. Even when real data is unlabeled, it remains useful to refine the activity pool, calibrate resident behavior statistics, identify sensor malfunctions from environmental factors, and analyze walking patterns based on estimated trajectories checked visually.

\section{Conclusion}

Our goal is the long-term detection of common anomalies among elderly individuals living alone using privacy-preserving sensors in smart homes. Using sensor data simulation, we detect six types of anomalies including falls while walking, falls while standing, wandering, forgetting to turn off home appliances and faucets, being housebound, and being semi-bedridden, using ambient sensors including infrared motion sensors, pressure sensors, door sensors, and cost sensors of home appliances and faucets. By utilizing simulated sensor data to train detection classifiers, we avoided the high costs associated with long-term data collection in the real world.

The need for a cost-effective solution to detect multiple anomalies led to developing a unified sensor preprocessing framework. This framework allows for detecting anomalies with various occurrence durations using a set of ambient sensors. Our detection classifiers are trained using simulation data with anomaly labels derived from predefined anomaly models. As the evaluation metrics, we considered anomalies correctly detected if the true time interval overlapped with the detected abnormal time interval. According to our experimental results using simlator-ganerated test data, this metric indicates that our detection methods correctly detect true (model-generated) anomalies with a sensitivity of over 0.9, and false alarms occur less than once every 50 days, except for fall while walking, where the sensitivity is 0.75. Though this result is difficult to compare related works because the experiment settings and performance metrics varies in existing studies, our methods performed competitively with existing techniques, especially for detecting falls and wandering. The ambient sensors based detection methods for forgetting, being housebound, and being semi-bedridden were newly proposed and evaluated using simulated data, achieving high precision and recall.

This approach demonstrates that assumptions about floor plans, sensor arrangements, and activity patterns can enable effective anomaly detection without relying on real data. This highlights the potential for real-time anomaly detection in smart home to enhance long-term elderly care. Future research will focus on refining simulation models and expanding their applicability to real-world scenarios.

\bibliographystyle{IEEEtran}  
\bibliography{mybib.bib}  

\end{document}